\documentclass{article}

\PassOptionsToPackage{numbers,sort&compress}{natbib}
\usepackage[final,sglblindworkshop]{neurips_2026}
\workshoptitle{I Can't Believe It's Not Better (ICBINB): Failure Modes of AI in Biology}

\usepackage[utf8]{inputenc}
\usepackage[T1]{fontenc}
\usepackage[hidelinks]{hyperref}  %
\usepackage{url}
\usepackage{booktabs}
\usepackage{amsfonts}
\usepackage{nicefrac}
\usepackage{microtype}
\usepackage{xcolor}
\usepackage{graphicx}
\usepackage{caption}
\usepackage{tikz}
\usetikzlibrary{arrows.meta,positioning,shapes.geometric}
\usepackage{pgfplots}
\pgfplotsset{compat=1.17}
\usepackage[most]{tcolorbox}
\usepackage{amssymb}
\usepackage{placeins}

\definecolor{accent}{RGB}{38,84,124}      %
\definecolor{highlight}{RGB}{191,111,7}   %
\definecolor{okgreen}{RGB}{35,120,70}     %
\definecolor{badred}{RGB}{214,65,58}      %

\definecolor{figred}{RGB}{214,65,58}      %
\definecolor{figblue}{RGB}{201,220,240}   %
\definecolor{figblueL}{RGB}{220,232,245}  %

\definecolor{set2teal}{RGB}{102,194,165}   %
\definecolor{set2orange}{RGB}{252,141,98}  %
\definecolor{set2blue}{RGB}{141,160,203}   %
\definecolor{set2pink}{RGB}{231,138,195}   %
\definecolor{set2green}{RGB}{166,216,84}   %
\definecolor{set2yellow}{RGB}{255,217,47}  %
\definecolor{set2tan}{RGB}{229,196,148}    %
\definecolor{set2gray}{RGB}{179,179,179}   %

\newsavebox{\diagmarkbox}
\sbox{\diagmarkbox}{\tikz[baseline=-0.6ex]
  \draw[accent, mark size=1.9pt, mark options={color=accent, fill=accent}]
  plot[mark=*] coordinates {(0,0)};}
\newsavebox{\rescmarkbox}
\sbox{\rescmarkbox}{\tikz[baseline=-0.6ex]
  \draw[okgreen, mark size=1.8pt, mark options={color=okgreen, fill=okgreen}]
  plot[mark=square*] coordinates {(0,0)};}

\newtcolorbox{findingsbox}{
  enhanced, colback=accent!4, colframe=accent, boxrule=0.7pt, arc=1.5mm,
  left=7pt, right=7pt, top=5pt, bottom=5pt,
  title=\textbf{Main findings}, fonttitle=\small, coltitle=white,
  colbacktitle=accent, breakable}
\newtcolorbox{takeawaybox}{
  enhanced, colback=highlight!6, colframe=highlight, boxrule=0.7pt, arc=1.5mm,
  left=7pt, right=7pt, top=5pt, bottom=5pt,
  title=\textbf{Takeaway}, fonttitle=\small, coltitle=white,
  colbacktitle=highlight, breakable}

\newtcolorbox{placeholder}[1][]{
  enhanced, colback=badred!5, colframe=badred, boxrule=1pt, arc=1mm,
  borderline={0.4pt}{2pt}{badred, dashed},
  left=7pt, right=7pt, top=5pt, bottom=5pt,
  title=\textbf{PLACEHOLDER --- NO DATA YET}, fonttitle=\small\bfseries,
  coltitle=white, colbacktitle=badred, breakable, #1}

\newtcolorbox{defbox}{
  enhanced, colback=gray!4, colframe=gray!45, boxrule=0.5pt, arc=1mm,
  left=6pt, right=6pt, top=4pt, bottom=4pt, breakable}

\definecolor{newblue}{RGB}{0,86,196}

\newcommand{\promptvar}[1]{$\langle$\textit{#1}$\rangle$}

\newcommand{\nextq}[2][Next question]{\textcolor{accent}{$\hookrightarrow$}~\textbf{\textcolor{accent}{#1}} --- \textit{#2}}
\newcommand{\takeaway}[1]{\textcolor{highlight}{$\blacktriangleright$}~\textbf{\textcolor{highlight}{Takeaway}} --- \textit{#1}}

\makeatletter
\newcommand{\blfootnote}[1]{%
  \begingroup
    \renewcommand{\thefootnote}{}%
    \renewcommand{\@makefnmark}{}%
    \footnotetext{#1}%
  \endgroup
}
\makeatother

\title{Treating the Symptom, Missing the Cause:\\[2pt]
  \large Anatomy of a Scientific Reasoning Failure}

\author{%
  Anna Gogleva\thanks{Lead and corresponding author.} \quad Thomas Charman \quad Martin Balla \quad Karl Gemayel \\
  Relation Therapeutics \\
  \texttt{anna.gogleva@relationrx.com}%
}

\begin{document}

\maketitle
\blfootnote{The benchmark environment is available at \url{https://github.com/RelationRx/luca} (CC BY-NC 4.0).}

\begin{abstract}
Scientific agents can find interventions that work without understanding why they do. We study this failure mode in controlled synthetic regulatory systems, where agents must experimentally identify a hidden fault and rescue the resulting phenotype. Across seven LLMs, every model restores the healthy phenotypes more often than it identifies the fault. Claude Opus 4.6 and Sonnet 4.6 rescue every organism while diagnosing only 55\% and 35\% of them. This failure is not simply due to insufficient exploration or unavailable evidence. In most cases, the cause is experimentally identifiable, and agents encounter evidence that distinguishes competing causes in 91\% of relevant sessions, and still frequently misdiagnose the system.

Targeted probes reveal where experimental reasoning breaks down. Models reliably interpret decisive evidence and revise hypotheses when the causal comparison is explicit, and can locate that evidence within a full experimental record. However, they struggle to predict the outcome of the experiment under each competing cause, a necessary step in using the observed result to distinguish between them. Agents can therefore gather the right evidence and still infer the wrong cause. As AI systems take on a growing role in synthesising scientific evidence, these findings highlight the need to evaluate how they use evidence to reach causal conclusions, alongside whether their interventions succeed.
\end{abstract}

\section{Introduction}
\label{sec:intro}

LLMs are moving from answering scientific questions to participating in the experimental process itself. They generate hypotheses \citep{lu2024aiscientist,majumder2024discoverybench,dobrowolska2026tinymoves,hypospace2025}, propose experiments \citep{laurent2024labbench}, interpret results \citep{wu2025perturbqa,zhao2025biomaze,chen2025scienceagentbench}, and decide what to test next \citep{boiko2023autonomous}. This shift makes the process by which an agent reaches a scientific conclusion increasingly important, not just the quality of its final output. An agent may reach a useful, even correct, result without understanding the system it is studying \citep{resultswithoutreasoning2026}.

In biology, one particularly important case is the distinction between successful intervention and causal understanding. An intervention can restore a phenotype without identifying its cause, just as a treatment can relieve symptoms without modifying the underlying disease mechanism. An agent judged only by its outputs can therefore appear scientifically competent while holding an incorrect causal model, or none at all \citep{medeinst2026,causalgame2026,causalab2026}.

This raises a harder question for evaluating scientific agents. When an agent reaches the wrong causal conclusion, where did the reasoning fail? It may explore too little, choose experiments that cannot distinguish competing explanations, encounter informative evidence but fail to recognise its significance, or fail to integrate evidence accumulated across the investigation. Alternatively, the available experiments may simply be unable to resolve the ambiguity. Outcome-based evaluation cannot distinguish these failure modes. Large-scale audits of scientific agents report symptoms of the same problem: evidence ignored in most traces, refutation rarely triggering revision \citep{resultswithoutreasoning2026}, but without ground truth they cannot locate its source.

Answering this question requires knowing more than whether the final answer is correct. For each reasoning session, we need to know which alternative hypotheses remain plausible, which experiments could distinguish them, what those experiments would reveal, and what evidence the agent actually encountered. Existing interactive benchmarks tend to evaluate the informativeness of chosen experiments prospectively \citep{boxinggym2025}, but generally cannot tell us what alternative experiments would have revealed. Fixed biological datasets provide even less freedom to ask what an agent could have discovered through a different experiment.

We therefore study agents in synthetic regulatory systems, where they can freely perturb an unknown network, observe its response, diagnose a hidden fault, and design an intervention to restore the healthy phenotype. The underlying mechanism is fully known to us but hidden from the agent. This gives us counterfactual access to every possible experiment. We can map the competing causes and the experiments that distinguish them, then trace what the agent actually tests and how it uses the evidence.

We begin from an apparent contradiction: \textbf{\textit{agents can fix systems they cannot explain}}. We then trace this gap through the experimental investigation. Diagnostic failures often persist even when the cause is experimentally identifiable and evidence distinguishing competing causes appears in the agent's own trajectory. Targeted probes further show that models can interpret such evidence, and locate it within a full experimental record, but cannot derive what phenotype readouts their candidate hypotheses predict based on it. Successful investigations also exhibit a distinctive contrastive signature, echoing experimental strategies used for causal attribution in biology.

Together, these results expose a failure mode that endpoint evaluation can hide: scientific agents may obtain the evidence required for causal inference without successfully carrying it through to the causal conclusion. Successful intervention is therefore not, by itself, evidence of causal understanding.

\section{A controlled environment for experimental discrimination}
\label{sec:setup}

\begin{figure}[!b]
\centering
\includegraphics[width=0.9\textwidth]{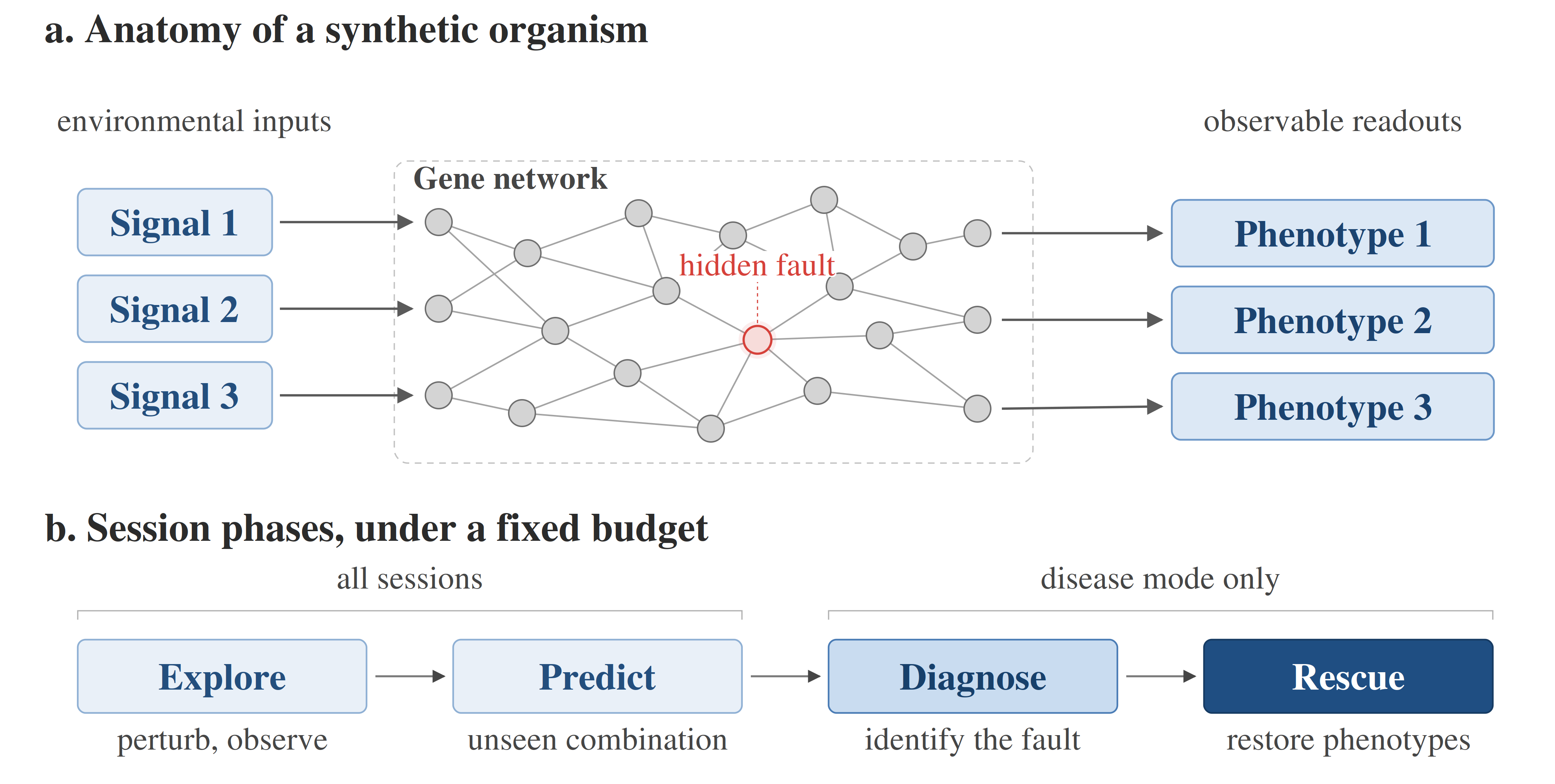}
\caption{\textbf{Controlled experimental environment.} \textbf{(a)} A synthetic organism maps environmental inputs through an unknown gene regulatory network to three observable phenotypes; one example hidden fault is highlighted. Agents cannot observe the network directly and must learn about the system through perturbations. \textbf{(b)} Each session proceeds under a fixed experimental budget. All sessions include exploration and prediction; disease sessions additionally require diagnosis of a hidden fault followed by an intervention to rescue the phenotypes. }
\label{fig:concept}
\end{figure}

We study agents in controlled synthetic regulatory systems with fully known
ground truth (see the LUCA environment \citep{gogleva2026luca}). Each
contains 20 genes connected in a regulatory network,
three external signals, and three observable phenotypes
(Figure~\ref{fig:concept}a). The systems are generated procedurally,
their dynamics are deterministic, and gene and signal names are
invented, so agents cannot rely on prior biological
knowledge and must learn through experimentation. The experimental budget
is sufficient to perturb every gene at least once, so an informative
experiment left unrun reflects the agent's choices rather than a tight
budget. The design follows synthetic networks used as controlled
ground-truth systems
\citep{vandenbulcke2006syntren,schaffter2011genenetweaver,tjarnberg2017genespider}.

Each system has a healthy state. In addition, we create a disease condition by
introducing one hidden fault: a gene fixed at high or low activity, or an
external signal permanently active.
\looseness=-1 Under a fixed budget, the agent can
increase~$\uparrow$, decrease~$\downarrow$, or block~$\varnothing$ genes,
activate signals, and observe the resulting phenotypes. It then predicts
the outcome of an unseen perturbation combination, commits to a diagnosis
of the faulty component and failure mode, and proposes a \emph{rescue}
intended to restore the healthy phenotypes (Figure~\ref{fig:concept}b).
Diagnosis must be submitted before rescue, so the rescue outcome cannot
revise it. For more details about the environment setup see Appendix~\ref{app:env}.

We score \emph{diagnosis} as correct only when the agent identifies both
the faulty component and its failure mode. \emph{Rescue} is correct when
the phenotypes return to their healthy range. These two outcomes can
dissociate, which is where our investigation begins.

\FloatBarrier
\section{Agents can fix systems they cannot diagnose}
\label{sec:gap}
\begin{figure}[!ht]
\centering
\begin{tikzpicture}[font=\scriptsize]

\node[font=\footnotesize\bfseries, anchor=north west] at (0,5.3)
  {(a) Network structure creates false leads};

\node[circle, draw=figred, fill=figred!12, minimum size=6.5mm,
      inner sep=0pt, font=\scriptsize] (A) at (3.2,4.35) {A};
\node[circle, draw=gray!55, fill=gray!18,
      minimum size=6.5mm, inner sep=0pt, font=\scriptsize] (X) at (3.2,3.08) {X};
\node[circle, draw=accent!70, fill=figblue,
      minimum size=6.5mm, inner sep=0pt, font=\scriptsize] (B) at (3.2,1.82) {B};
\node[rounded corners=2pt, draw=accent!55, fill=figblueL, text=accent,
      inner sep=3pt, minimum height=6.5mm, font=\scriptsize] (P) at (3.2,0.55)
      {abnormal phenotype};

\draw[-{Stealth[length=1.5mm]}, gray!60] (A) -- (X);
\draw[-{Stealth[length=1.5mm]}, gray!60] (X) -- (B);
\draw[-{Stealth[length=1.5mm]}, gray!60] (B) -- (P);

\node[font=\scriptsize, figred, anchor=east] at (2.55,4.35)
  {true cause};
\node[font=\scriptsize, accent, anchor=east, align=center] at (2.55,1.82)
  {downstream gene\\\textit{a plausible false lead}};

\node[font=\footnotesize\bfseries, anchor=north west] at (6.4,5.3)
  {(b) The diagnosis--rescue gap};
\begin{axis}[
  at={(6.4cm,0.15cm)}, anchor=south west,
  width=6.0cm, height=4.5cm, scale only axis,
  xlabel={Diagnosis \usebox{\diagmarkbox} / Rescue
          \usebox{\rescmarkbox} score}, xmin=0, xmax=1.05,
  symbolic y coords={Random,CausalSweep,Heuristic,gap,GPT-4o,GPT-5.1,
                     DeepSeek V3,Qwen 3.5,GPT-4.1,Sonnet 4.6,Opus 4.6},
  ytick=data, y tick label style={font=\scriptsize},
  tick label style={font=\scriptsize}, label style={font=\footnotesize},
  xmajorgrids, grid style={gray!15},
  clip=false, %
]
\draw[gray!50, dotted, line width=0.6pt] (axis cs:0,gap) -- (axis cs:1.05,gap);
\node[font=\scriptsize\itshape, gray, anchor=east, inner xsep=2pt]
  at (axis cs:0,gap) {baselines};
\draw[gray!55, line width=0.7pt] (axis cs:0.075,Random)         -- (axis cs:0.200,Random);
\draw[gray!55, line width=0.7pt] (axis cs:0.100,CausalSweep)    -- (axis cs:0.650,CausalSweep);
\draw[gray!55, line width=0.7pt] (axis cs:0.100,Heuristic)      -- (axis cs:0.700,Heuristic);
\draw[gray!55, line width=0.7pt] (axis cs:0.075,GPT-4o)         -- (axis cs:0.100,GPT-4o);
\draw[gray!55, line width=0.7pt] (axis cs:0.150,GPT-5.1)        -- (axis cs:0.400,GPT-5.1);
\draw[gray!55, line width=0.7pt] (axis cs:0.200,DeepSeek V3)    -- (axis cs:0.900,DeepSeek V3);
\draw[gray!55, line width=0.7pt] (axis cs:0.200,Qwen 3.5)       -- (axis cs:0.900,Qwen 3.5);
\draw[gray!55, line width=0.7pt] (axis cs:0.250,GPT-4.1)        -- (axis cs:0.650,GPT-4.1);
\draw[gray!55, line width=0.7pt] (axis cs:0.350,Sonnet 4.6)     -- (axis cs:1.000,Sonnet 4.6);
\draw[gray!55, line width=0.7pt] (axis cs:0.550,Opus 4.6)       -- (axis cs:1.000,Opus 4.6);
\addplot[only marks, mark=*, mark size=1.9pt, accent] coordinates {
  (0.150,GPT-5.1) (0.250,GPT-4.1) (0.075,Random) (0.075,GPT-4o)
  (0.100,Heuristic) (0.200,DeepSeek V3) (0.100,CausalSweep)
  (0.200,Qwen 3.5) (0.350,Sonnet 4.6) (0.550,Opus 4.6)};
\addplot[only marks, mark=square*, mark size=1.8pt, okgreen] coordinates {
  (0.400,GPT-5.1) (0.650,GPT-4.1) (0.200,Random) (0.100,GPT-4o)
  (0.700,Heuristic) (0.900,DeepSeek V3) (0.650,CausalSweep)
  (0.900,Qwen 3.5) (1.000,Sonnet 4.6) (1.000,Opus 4.6)};
\end{axis}
\end{tikzpicture}
\caption{\textbf{The diagnosis--rescue gap.} \textbf{(a)} A downstream
gene can provide a plausible diagnosis and rescue target, restoring the
phenotype without correcting the true fault. \textbf{(b)} Diagnosis and rescue at medium difficulty (20 seeds)
for seven LLMs and three non-LLM baselines. Lines connect the two scores
for each agent. CIs and prediction accuracy are in
Appendix~\ref{app:gap-ci} (Table~\ref{tab:gap-ci}).}
\label{fig:gap}
\end{figure}
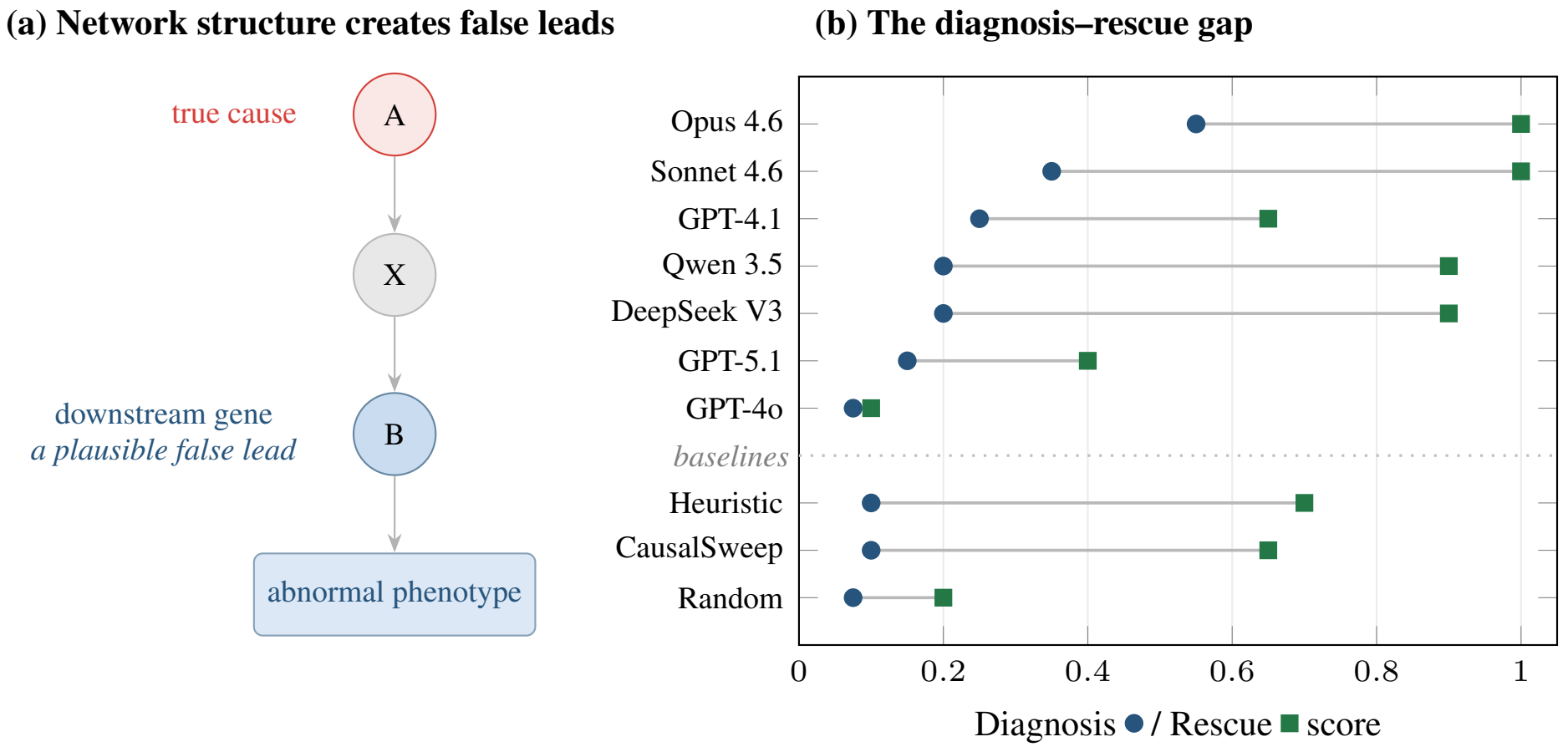

\looseness=-1
Successful intervention does not necessarily imply correct causal understanding. Similar dissociations between task success and causal reasoning have been observed in other interactive settings
\citep{causalgame2026,causalab2026,task2quiz2026}. We reproduce it here:
agents are consistently better at restoring a system to its healthy
state than at identifying the fault that caused it
(Figure~\ref{fig:gap}). On medium-difficulty organisms, Claude Opus 4.6
rescues all 20 organisms but correctly diagnoses only 55\%; Sonnet 4.6 likewise
achieves perfect rescue but only 35\% diagnosis. Rescue scores exceed diagnosis
for every model we evaluate, though the two confidence intervals overlap
for the weakest performers (Appendix~\ref{app:gap-ci}).

Compensatory effects inherently present in networks could explain this gap. A downstream gene can be an effective point of intervention without being the source of the fault (Figure~\ref{fig:gap}a). Changing it may restore the phenotypes by compensating for the fault, while leaving its cause unidentified. So, even in these small networks, finding an intervention and identifying the causal fault are distinct problems that can be non-trivial. Agents are explicitly asked to locate the causal gene or signal (Appendix~\ref{app:prompts}).

\section{The investigation}
\label{sec:investigation}

The diagnosis--rescue gap itself is not our main finding. Our controlled environment lets us investigate why it occurs by tracing the agents' experimental trajectories and testing possible explanations. The non-LLM baselines do not log trajectories, so the behavioural analyses that follow cover the seven LLMs (420 sessions). We begin with the simplest hypothesis.
\nopagebreak[4]

\nextq[First question]{Perhaps agents misdiagnose because they do not investigate the system thoroughly enough?}

\subsection{First suspect: insufficient exploration}
\label{sec:explore}

\begin{figure}[!ht]
\centering
\includegraphics{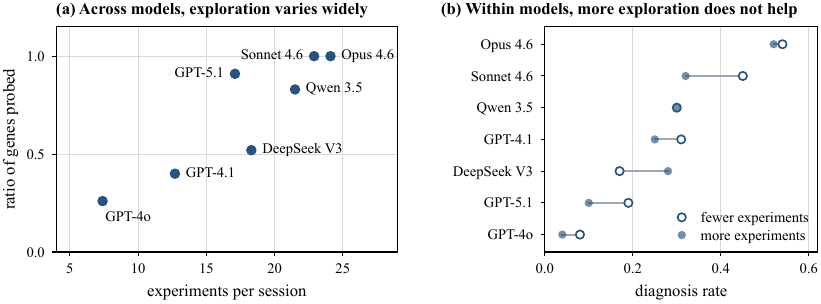}
\caption{\textbf{Exploration does not simply explain diagnostic success.}
\textbf{(a)} Models differ substantially in both the number of experiments they run and the ratio of genes they probe. \textbf{(b)} Within each model, sessions above versus below its median experiment count show no consistent advantage from greater exploration.
}
\label{fig:exploration}
\end{figure}

If insufficient exploration explains the gap, agents that experiment
longer or more broadly should diagnose more accurately. At first glance this seems plausible. Models investigate the same organisms
very differently (Figure~\ref{fig:exploration}a): the average number of
experiments per session ranges from 7.4 to 24.1, while the ratio of genes
probed ranges from about a quarter of the genes in a network to all of them. Models that explore more also tend to diagnose better. 

Within each model, however, the pattern disappears
(Figure~\ref{fig:exploration}b). Sessions above each model's median
experiment count have a diagnosis rate of 0.26, compared with 0.29 for
sessions below it. Experiment count is likewise uncorrelated with
diagnosis within models ($r=-0.05$). Breadth shows the same pattern: among
the five models with varying coverage, the mean within-model correlation
between gene coverage and diagnosis is $+0.04$. Claude Opus 4.6 and
Sonnet 4.6 probe every gene in every session, leaving no within-model
variation to compare.

\takeaway{Diagnostic failure cannot be explained simply by agents
running too few experiments or probing too few genes.}

\nextq{Perhaps the problem is not insufficient exploration, but a lack of experiments capable of identifying the cause?}
\subsection{Second suspect: missing evidence}
\label{sec:evidence}

More exploration did not explain diagnostic success. This raises a
more basic possibility: perhaps some organisms simply cannot be diagnosed
from the experiments available to the agent? If two different faults
produce the same observations under every permitted intervention, no
experimental strategy can tell them apart.

\begin{figure}[!ht]
\centering
\begin{tikzpicture}[
  x=0.93cm,   %
  font=\scriptsize,
  every node/.style={align=center},
  obs/.style={draw=gray!65, fill=gray!10, rounded corners=2pt, inner sep=3.5pt},
  hyp/.style={draw=accent, fill=accent!8, rounded corners=2pt, inner sep=3.5pt},
  res/.style={draw=figred, fill=figred!8, rounded corners=2pt, inner sep=3.5pt},
  tru/.style={draw=okgreen, fill=okgreen!8, rounded corners=2pt, inner sep=3.5pt},
  note/.style={font=\scriptsize\itshape, text=gray!35!black, fill=white,
               inner sep=2pt},
  ln/.style={draw=gray!60!black},
]

\node[font=\scriptsize\bfseries] at (0,1.35) {(a) Two faults fit the observation};
\node[obs] (O) at (0,0.7)  {low phenotype};
\node[hyp] (A) at (-1.9,-0.35)  {hypothesis A:\\gene A is the cause};
\node[hyp] (B) at (1.9,-0.35)   {hypothesis B:\\gene B is the cause};
\draw[ln, -{Stealth[length=1.4mm]}] (A) -- (O);
\draw[ln, -{Stealth[length=1.4mm]}] (B) -- (O);

\draw[->, thick, gray!45!black] (3.6,0.28) -- (4.9,0.28);
\node[font=\scriptsize, text=gray!35!black] at (4.25,0.62) {discriminating experiment};

\node[font=\scriptsize\bfseries] at (8.5,1.35) {(b) One experiment separates them};
\node[obs] (P) at (8.5,0.7)   {perturb gene A};
\node[res] (RA) at (6.6,-0.35)     {if A is the cause:\\phenotype = 0.8};
\node[res] (RB) at (10.4,-0.35)    {if B is the cause:\\phenotype = 0.4};
\draw[ln, -{Stealth[length=1.4mm]}] (P) -- (RA);
\draw[ln, -{Stealth[length=1.4mm]}] (P) -- (RB);

\draw[gray!30] (-2.9,-1.05) -- (11.5,-1.05);
\node[font=\scriptsize\bfseries, anchor=west] at (-2.9,-1.4)
  {(c) Can the true fault be identified?};

\node[obs] (T0) at (4.3,-1.6) {initial observation};
\node[font=\scriptsize, text=gray!35!black] (Q1) at (4.3,-2.35)
  {alternative fault?};
\node[tru] (L1) at (1.0,-3.15) {unambiguous\\(14 organisms)};
\node[font=\scriptsize, text=gray!35!black] (Q2) at (7.6,-3.15)
  {experiment separates?};
\node[tru] (L2) at (5.0,-4.05) {resolvable ambiguity\\(39 organisms)};
\node[res] (L3) at (10.2,-4.05) {structural ambiguity\\(7 organisms)};

\draw[ln] (T0) -- (Q1);
\draw[ln] (Q1) -- (L1) node[midway, above=1pt, font=\tiny\itshape,
  text=gray!45!black] {no};
\draw[ln] (Q1) -- (Q2) node[midway, above=1pt, font=\tiny\itshape,
  text=gray!45!black] {yes};
\draw[ln] (Q2) -- (L2) node[midway, left=2pt, font=\tiny\itshape,
  text=gray!45!black] {yes};
\draw[ln] (Q2) -- (L3) node[midway, right=2pt, font=\tiny\itshape,
  text=gray!45!black] {no};
\end{tikzpicture}
\caption{\textbf{Alternative hypotheses and discriminating experiments.}
\textbf{(a)} Two different faults can produce the same abnormal
readout, leaving both as plausible causes. \textbf{(b)} An experiment is
\emph{discriminating} if its outcome differs depending on which fault is
present. Values in (a) and (b) are illustrative. \textbf{(c)} We classify
the 60 organisms by whether an alternative fault fits the initial
observation and, if so, whether an available experiment can distinguish
it from the true fault.}
\label{fig:alternatives}
\end{figure}
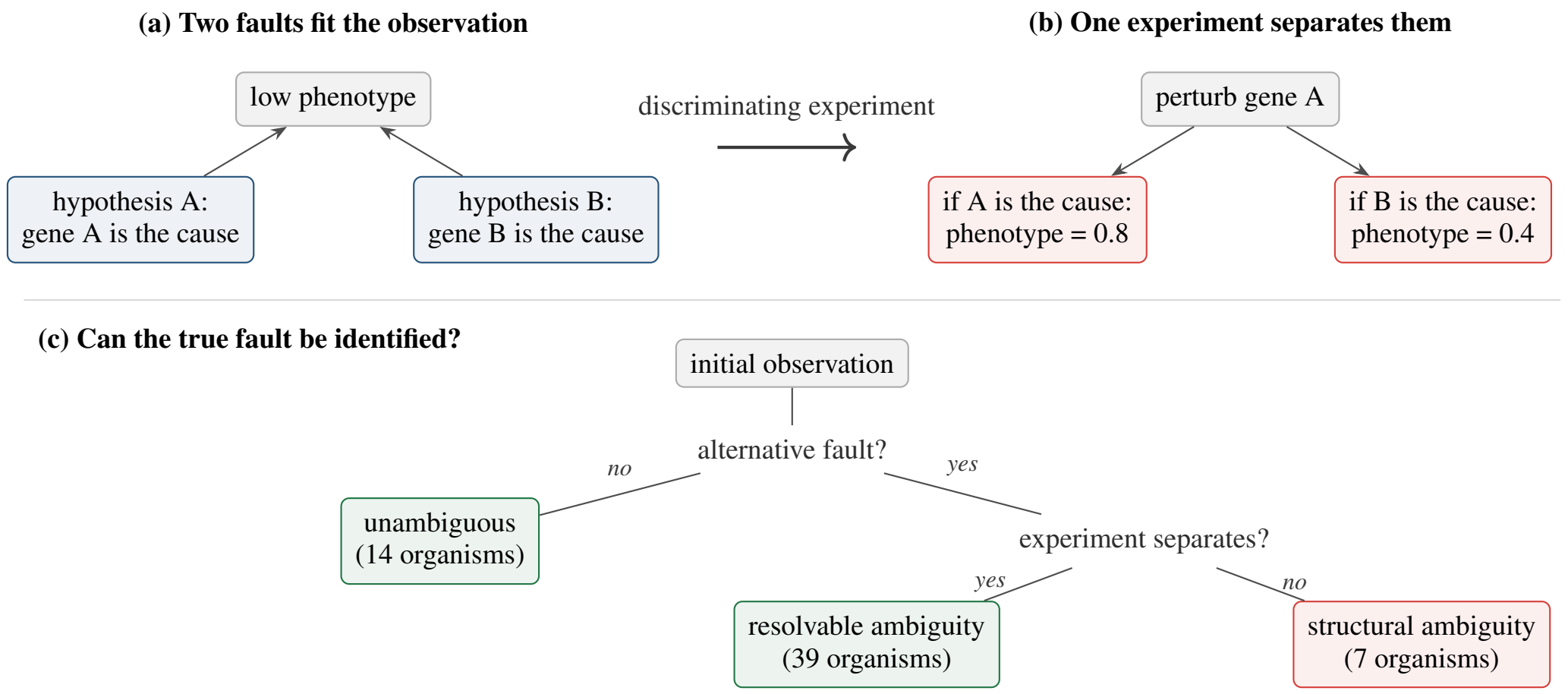

Our controlled setting lets us check this directly. Suppose two faults produce the same abnormal phenotype: from that observation alone, both are plausible causes. We call the second an \emph{alternative hypothesis}. An experiment is \emph{discriminating} if its outcome would differ depending on which fault is present. Because the underlying system is known, we can simulate every possible fault and intervention, including experiments the agent never ran (Figure~\ref{fig:alternatives}).

This separates the 60 organisms into three cases. In 14, only the true fault fits the initial phenotype (\emph{unambiguous}). In 39, alternatives initially fit but available experiments can distinguish them from the truth (\emph{resolvable ambiguity}). Only 7 contain an alternative that no available experiment can distinguish from the true fault (\emph{structural ambiguity}). So, in 53 of 60 organisms, the cause is identifiable using the evidence available to the agent.

The mere presence of competing hypotheses does not explain the failures
either. Diagnosis is similar on unambiguous and resolvably ambiguous
organisms: 0.31 [0.16, 0.46] versus 0.30 [0.23, 0.37], respectively
(bootstrap 95\% CIs clustered by organism).

\takeaway{Agents had access to experiments capable of identifying the
cause in most organisms; poor diagnosis cannot be attributed to
missing information.}

\nextq{If the evidence was available, do agents actually seek it?}

\subsection{The culprit: unused evidence}
\label{sec:discriminate}

\begin{figure}[!ht]
\centering
\includegraphics{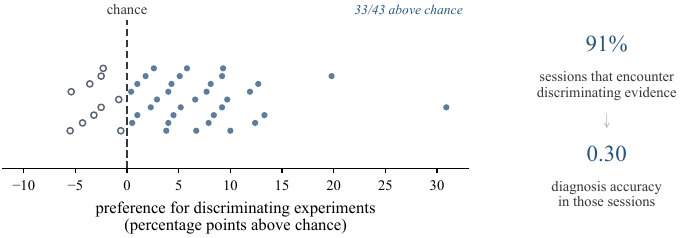}
\caption{\looseness=-1 \textbf{Agents usually encounter discriminating evidence, but barely use it.} Each point shows, for one organism with a resolvable
alternative ($n=43$), how often agents choose discriminating experiments
relative to random choice. Agents show only a small preference above chance
(24.1\% vs.\ 19.2\%). Nevertheless, 91\% of sessions encounter
discriminating evidence before diagnosis, yet diagnosis accuracy in
those sessions is only 0.30.}
\label{fig:seek}
\end{figure}

For each organism with at least one resolvable alternative ($n=43$), we
\looseness=-1
compare the experiments agents choose with uniform random selection from
the same available interventions. Across 6{,}596 experiments run before agents commit
their diagnosis, 24.1\% of agents'
choices discriminate between the true fault and an alternative, compared
with 19.2\% under random selection (Figure~\ref{fig:seek}). The difference
is systematic across organisms ($t=4.6$) but small: agents show only a
weak preference for experiments that separate competing hypotheses,
echoing the confirmation bias reported in LLMs
\citep{failingtofalsify2026}.

This comparison is an oracle audit: discriminating experiments are identified using the ground truth and dynamics that are, by design, not available to the agents. Still, agents can test competing hypotheses by choosing experiments they expect to produce different outcomes depending on which hypothesis is
correct.

Despite this weak preference, agents usually encounter discriminating
evidence anyway. In 91\% of committed sessions on these organisms, they
run at least one
discriminating experiment before diagnosis, yet diagnosis accuracy in
these sessions is only 0.30. The evidence is rarely difficult to isolate:
97\% of these sessions contain a discriminating result from a single
perturbation. Nor does it arrive too late: agents run a median of 16
further experiments after first encountering it. So, the evidence is
usually present in the agent's own trajectory, in a simple form and well
before it commits.

\begin{figure}[!ht]
\centering
\includegraphics{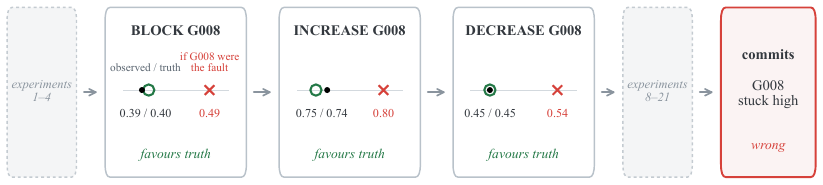}
\caption{\textbf{One agent holds the evidence and commits against it.}
Experiments 5--7 probe the agent's eventual diagnosis, G008, in three
directions. In each case, the observed readout of one phenotype (black)
matches the true fault's prediction (green) rather than G008's (red). The agent nevertheless
commits to G008 14 experiments later.}
\label{fig:trace}
\end{figure}

Figure~\ref{fig:trace} shows one example. The agent considers G008 as a
possible cause and tests it in three directions. Each result favours the
true fault over G008, yet the agent continues for another 14 experiments
and eventually diagnoses G008. Its subsequent rescue succeeds despite
the wrong diagnosis. Here, the problem is not obtaining evidence against
the eventual diagnosis, but carrying that evidence forward into the
causal conclusion.

\takeaway{Agents choose discriminating experiments barely more often
than chance, but encounter them in nearly every session. Most still
reach the wrong diagnosis.}

\nextq{Why does evidence that distinguishes competing causes fail to
change the diagnosis?}

\subsection{The signature: successful diagnosis is contrastive}
\label{sec:bidirectional}

We first look for a behavioural signature of the
investigations that succeed. One pattern stands out: correct diagnosis is
strongly associated with perturbing the true causal component in both
directions before committing.

\begin{table}[!ht]
\centering
\caption{\textbf{Diagnosis accuracy by how agents perturb the true
cause}, with session counts in brackets. Testing in both directions is
associated with higher accuracy for every model. Five cells contain fewer
than ten sessions, so individual estimates should be interpreted
cautiously.}
\label{tab:bidirectional}
\footnotesize
\newlength{\dbarw}\settowidth{\dbarw}{\textbf{0.00}\,(156)}
\newlength{\dscw}\settowidth{\dscw}{\textbf{0.00}}
\newlength{\dctw}\settowidth{\dctw}{(156)}
\newlength{\dfw}
\newcommand{\dbg}[1]{%
  \setlength{\dfw}{#1\dbarw}%
  \ifdim\dfw>0pt\ifdim\dfw<2.8pt\setlength{\dfw}{2.8pt}\fi\fi%
  \rlap{\color{accent!10}\rule[-1.8pt]{\dbarw}{7.2pt}}%
  \rlap{\color{accent!50}\rule[-1.8pt]{\dfw}{7.2pt}}}
\newcommand{\dc}[2]{\dbg{#1}\makebox[\dscw][r]{#1}\,\makebox[\dctw][l]{(#2)}}
\newcommand{\db}[2]{\dbg{#1}\makebox[\dscw][r]{\textbf{#1}}\,\makebox[\dctw][l]{(#2)}}
{\renewcommand{\arraystretch}{1.05}%
\begin{tabular}{lccc}
\toprule
 & \multicolumn{3}{c}{true cause perturbed} \\
\cmidrule(lr){2-4}
model & \multicolumn{1}{c}{never} & \multicolumn{1}{c}{one direction} & \multicolumn{1}{c}{both directions} \\
\midrule
Opus 4.6      & \dc{0.00}{3} & \dc{0.00}{19} & \db{0.84}{38} \\
GPT-4.1       & \dc{0.00}{25} & \dc{0.06}{16} & \db{0.84}{19} \\
Sonnet 4.6    & \dc{0.00}{2} & \dc{0.00}{24} & \db{0.71}{34} \\
Qwen 3.5      & \dc{0.00}{8} & \dc{0.13}{30} & \db{0.64}{22} \\
GPT-4o        & \dc{0.00}{43} & \dc{0.08}{12} & \db{0.60}{5} \\
DeepSeek V3   & \dc{0.00}{21} & \dc{0.00}{16} & \db{0.57}{23} \\
GPT-5.1       & \dc{0.00}{6} & \dc{0.03}{39} & \db{0.53}{15} \\
\midrule
pooled        & \dc{0.00}{108} & \dc{0.04}{156} & \db{0.71}{156} \\
\bottomrule
\end{tabular}}
\end{table}

When agents perturb the true cause in only one direction, diagnosis
accuracy is 0.04; when they perturb it in both directions, accuracy rises
to 0.71 (Table~\ref{tab:bidirectional}). No agent correctly diagnoses a
cause it never perturbs (108 sessions). The same ordering holds for
every model: accuracy in the both-directions group ranges from 0.53 for
GPT-5.1 to 0.84 for Claude Opus 4.6.

Testing in both directions provides a contrast around the same
candidate cause: the agent observes how the system responds to opposing
changes rather than to a single intervention. This resembles the
complementary use of loss- and gain-of-function perturbations in
functional genomics, including CRISPRi and CRISPRa screens
\citep{gilbert2014crispria}. Successful diagnosis is therefore associated
not simply with more evidence, but with contrastive evidence about the
same candidate cause.

The association does not establish why this pattern predicts success.
Testing in both directions may make the causal evidence easier to
interpret, or agents that already suspect the correct cause may simply
be more likely to test it systematically.

\takeaway{Successful diagnosis has a strong contrastive signature:
agents that test the true cause in both directions are far more likely
to identify it correctly.}

\nextq{Can agents recognise and act on discriminating evidence when it is
put directly in front of them?}

\subsection{Probing where evidence use breaks down}
\label{sec:probes}

Agents usually encounter discriminating evidence but often fail to act
on it. To find out why, we replay real investigations and gradually expose evidence to the agents. For each of the 43 organisms with
a resolvable alternative we take one real session and cut it just
before the agent committed a diagnosis. 42 of the 43 records come from
Opus~4.6 sessions and one from a Sonnet~4.6 session, with a median of
22 experiments and 1--1.5k tokens of log. All seven models read the
same experimental session records. 

\begin{figure}[!ht]
\centering
\includegraphics{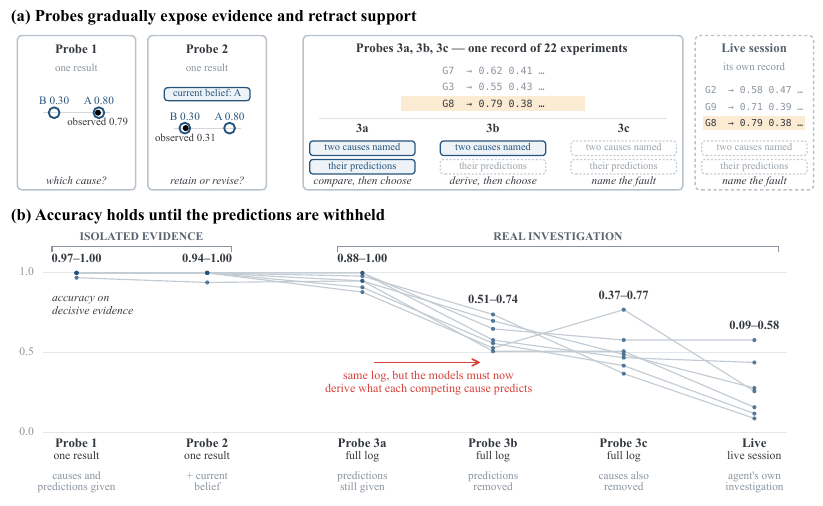}
\caption{\textbf{Evidence use breaks in context.}
\textbf{(a)} Each probe shows more of the investigation and
gives less help with it: the two competing causes, the readout each one
predicts, both, or neither. \textbf{(b)} Accuracy on decisive
evidence, each gray line represents a model; per-model values
are in Tables~\ref{tab:probes12}
and~\ref{tab:probe3}. Accuracy stays close to 1.00 for
as long as the predicted readouts are given (Probes~1--2 and 3a), falls
once the models must work them out (3b--c), and falls further in live
sessions.}
\label{fig:evidence-use}
\end{figure}

We designed three probe types, following increasing difficulty of the task. Probe~1 shows a single experiment; probe~2 adds a working hypothesis to defend or update; probes~3a to~3c show the whole log and
withdraw the support in even more granular increments
(Figure~\ref{fig:evidence-use}). Appendix~\ref{app:probes} provides
construction details for each probe, together with confidence intervals
and a repeat run.

\vspace{-12pt}

\paragraph{Isolated evidence is easy to interpret.}
Probe~1 provides two candidate causes, their predicted outcomes, and an
observed result; Probe~2 additionally assigns a current hypothesis and
asks whether to retain or revise it. On 69 discriminating results, six
of seven models are at or near ceiling on both tasks. DeepSeek~V3 is the
exception: it revises a contradicted hypothesis almost perfectly (0.94)
but retains a supported one in only 0.74 of cases. Thus, when the causal
comparison is explicit, models can both interpret decisive evidence and
revise their hypothesis.

Models are less reliable at recognising \emph{non-evidence}.
On 43 items, one per organism, where the two causes predict
readouts less than the measurement noise apart, Probe~1 accuracy falls to 0.30--0.53. Models usually
choose the numerically closer prediction rather than conclude that the
result is uninformative; in Probe~2, this triggers unwarranted revision
in 5--13 of 43 cases.

\paragraph{The same evidence becomes difficult to interpret
in the context of an experiment.}
Probe~3
has three layers, and each removes one kind of support
while the record itself stays the same. The first layer (3a) supplies
the two competing causes and the readout each one predicts for every
logged experiment, so the model only has to find the experiment that
separates them. Accuracy in this case is 0.88--1.00, matching the isolated-evidence
probes, although the decisive result stays buried where the source
agent left it. The second layer (3b) withholds those predictions, so
the model must work out for itself what each cause implies: accuracy
falls to 0.51--0.74. Performance of every model drops, and the two intervals do not
overlap for any of them. Finally, the third layer (3c) withholds the causes as
well, which leaves the live diagnosis question asked of a record the
model did not have to collect. Here accuracy reaches 0.37--0.77. These results suggest that the failure mode cannot be localised to finding the
decisive result in the experimental log. Repeating the first two
layers on records written by a different model reproduces the pattern,
with the same median gain of 0.35.

Reading the same record also compresses the large differences between models. On
these 43 organisms, live diagnosis ranges from 0.09 to 0.58; given
the identical experimental record, it ranges from 0.37 to 0.77. Live
performance therefore reflects both reasoning over evidence and the
evidence produced by the model's own investigation.

\takeaway{Models can interpret decisive evidence, revise a
hypothesis, and locate the decisive observation inside a full
experimental record. Even given appropriate and sufficient evidence, models
struggle to predict the phenotype readouts that competing explanations
of the malfunction would produce.}

\section{Conclusion}

\begin{figure}[!ht]
\centering
\includegraphics{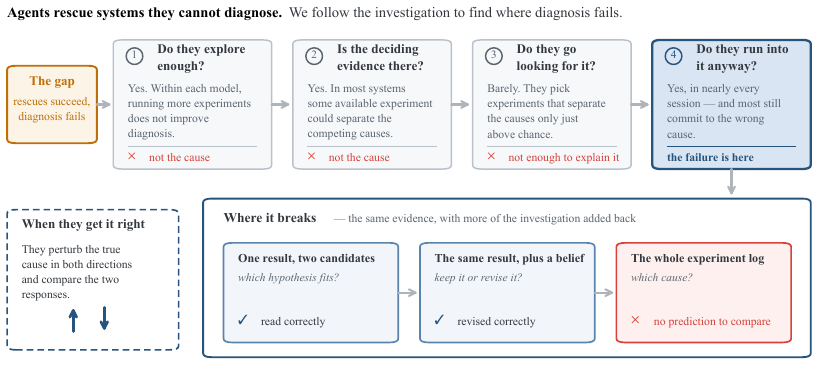}
\caption{\textbf{Tracing the diagnosis--rescue gap through the investigation.}
Successive analyses rule out insufficient exploration and, for most
systems, missing evidence, while bidirectional testing (dashed box) marks a correlate
of successful diagnosis. Probes localise the remaining failure:
models interpret and revise on isolated evidence, and still
locate it inside a full record, but cannot derive what each competing
cause predicts.}
\label{fig:summary}
\end{figure}

Scientific agents can restore a system's behaviour without identifying the underlying cause. We used a controlled experimental environment to ask where this gap comes from,
following the failure mode from experimental exploration to the use of evidence. Figure~\ref{fig:summary} shows the investigation as a chain: each step eliminates one explanation and hands its question to the next.

The failure proved surprisingly difficult to localise. Agents did not
simply explore too little. We established that discriminating evidence was usually accessible and appeared in most experimental sessions. We used targeted probes to dissect the reasoning process: agents interpreted isolated evidence and revised their hypotheses reliably, and recovered the decisive observation from a full experimental record once they were told what each competing cause predicted, but could not derive those predictions themselves. One possible explanation is that deriving those readouts requires a causal model of the organism, which the models would have to build from the record in front of them. Successful investigations nevertheless showed one distinctive signature: the true cause was tested in both directions far more often than in failed investigations.

The broader conclusion is that successful intervention, informative
experiments, and even access to decisive evidence do not by
themselves demonstrate causal understanding. Evaluating scientific agents
therefore requires looking inside the reasoning traces: what hypotheses
remain plausible, what evidence could distinguish them, what evidence the
agent obtains, and what it does with it. Our synthetic systems make these questions measurable by construction. They do not establish how often the same failures occur in real biology, but allow us to locate failures in scientific reasoning before asking whether they persist in more realistic settings.

\bibliographystyle{unsrt}
\bibliography{references}

\section*{Disclosure of LLM use}
Large language models are the object of study in this work: the seven
models evaluated throughout the paper are listed in
Appendix~\ref{app:explore}. Separately, the authors used an LLM-based
coding assistant while preparing this manuscript. Its role was limited to
editing the \LaTeX{} source, writing the plotting scripts that render
several of the figures from tabulated results, cross-checking reported
numbers and citations against the analysis outputs, and searching for
related work. It did not run experiments, choose analyses, or score any
model output: all benchmark and probe scoring is programmatic and uses
exact or graded string matching against ground truth
(Appendix~\ref{app:env}), with no model acting as a judge. The authors
designed the study, made every analysis decision, and take
responsibility for all claims in the paper.

\section*{Reproducibility statement}
The benchmark environment is available at
\url{https://github.com/RelationRx/luca}. Organisms are fully determined by a (seed, tier, $N$) triple and
generation is frozen at a tagged commit, so every system reported here can
be regenerated exactly; Appendix~\ref{app:env} gives the dynamics, the
fault model, the budget rule, and the scoring definitions. The behavioural
results rest on 420 sessions, one per (model, organism) pair over seven
LLMs and 60 disease organisms drawn from three sweeps. The discrimination
census that defines alternatives and discriminating sets is specified in
Appendix~\ref{app:census}, the exploration axes in
Appendix~\ref{app:explore}, and the construction, item counts, and
response budgets of all three probes in Appendix~\ref{app:probes}. The
task instructions and every probe prompt are reproduced verbatim in
Appendix~\ref{app:prompts}. Uncertainty is reported three ways: $t$-distribution
intervals over seeds for the per-model scores in
Appendix~\ref{app:gap-ci}, bootstrap intervals clustered by organism for
organism-level quantities, and Wilson intervals for item accuracies; a full second run of the isolated-evidence
probes is summarised in Appendix~\ref{app:probes}.

\section*{Ethics statement}
This work uses only synthetic regulatory systems generated by code. It
involves no human or animal subjects, no personal or clinical data, and no
experiments on biological material. The organisms are invented
abstractions rather than models of any real organism, and all gene,
signal, and phenotype names are constructed words with no correspondence
to real biological entities, so the paper conveys no information about
real pathways or interventions. The study evaluates the limitations of AI
systems on this synthetic task; we make no claim that the reported failure
rates transfer to real biological discovery, and we caution against
reading the diagnostic scores as evidence about any specific deployed
system.

\newpage
\appendix

\begin{center}
{\Large\bfseries Appendix}
\end{center}
\vspace{4pt}

\section{The LUCA environment}
\label{app:env}

This appendix summarises the benchmark environment and its validation.
The full design, generation pipeline, and validation experiments are in
the benchmark paper \citep{gogleva2026luca}; everything the results in
this paper depend on is restated here.

\paragraph{Organisms.}
An organism is a synthetic gene regulatory network of $N$ genes in
functional modules, driven by external signals and read out through
observable phenotypes. At $N{=}20$, the setting used throughout this
paper, an organism has 4 modules, 3 signals, and 3 phenotypes. Gene
states $s_i \in [0,1]$ evolve by a sigmoid update,
$s_i \leftarrow \sigma\!\big(\beta(\textstyle\sum_j w_{ij} s_j + c_i +
b_i)\big) + \varepsilon_i$, where positive weights activate and
negative weights inhibit, $c_i$ sums the active signal contributions,
$b_i$ is a gene bias, $\beta$ sets the steepness, and
$\varepsilon_i \sim \mathcal{N}(0, \sigma^2)$ is per-gene noise; the
network is iterated to a fixed point. Difficulty tiers (easy, medium,
hard) vary edge density, feed-forward and feedback structure, sigmoid
steepness, noise, and the fraction of latent genes; the medium tier has
no latent genes and moderate feedback. Every organism is fully
determined by a (seed, tier, $N$) triple.

\paragraph{Invented vocabulary.}
All entity names are constructed words with no correspondence to real
biological entities (genes \texttt{G001}--\texttt{G020}, signals such
as \texttt{burst-Y}, phenotypes such as \texttt{adaptation-score}),
removing the retrieval confound: an agent cannot look the system up, it
can only experiment on it. The benchmark paper \citep{gogleva2026luca}
tests this choice directly in a paired ablation (invented versus real human gene
names, crossed with prompt stance and difficulty; 320 sessions) and
finds no effect of the vocabulary in any paired contrast.

\paragraph{Faults, sessions, and scoring.}
In disease mode one gene or signal is locked into an abnormal state:
a gene locked high, a gene locked low, or a chronically active external
signal. The fault drives at least one phenotype outside its healthy
range, is never a latent gene, and by construction at least one
corrective single-target intervention exists. A session gives the agent
a fixed budget of $B = N + N_{\mathrm{sig}} + 8$ experiments (31 at
$N{=}20$) --- enough for a complete single-gene sweep with little room
for follow-up. The agent explores freely, answers two held-out
prediction tasks, commits to a single malfunction hypothesis, and only
then may test rescues: rescue is locked until the diagnosis is
committed, so a rescue can never be found by trial and error and then
rationalised into a diagnosis. Scoring: \emph{prediction accuracy} is
the fraction of phenotypes predicted within $\pm 0.10$ of ground truth;
\emph{diagnosis} is graded 1.0 (correct target and fault type), 0.5
(correct target, wrong type), 0.0 (wrong target); \emph{rescue} is the
fraction of aberrant phenotypes restored to the healthy range.
A reviewer of this paper found that the released scorer
instead counted every phenotype inside its healthy range, so an
intervention that changed nothing scored $2/3$ on an organism with one
aberrant phenotype of three. We thank them for the catch. The scorer
has been corrected and every rescue score reported here is computed on
the corrected metric; the benchmark release carries the fix. A
no-intervention control under the corrected metric scores
$0.150 \pm 0.171$ rather than zero, because in 3 of the 20
medium-difficulty organisms the aberrant phenotype lies within 0.02 of
its healthy range and simulation noise alone can move it inside
(Appendix~\ref{app:gap-ci}). The environment itself is
deterministic: the three non-LLM baselines reproduce the scores
reported here exactly from the released code. The seven LLM rows are
single sampled runs, so they cannot be reproduced exactly, by us or by
anyone else.
Section~\ref{sec:gap} uses the graded diagnosis score; the
investigation sections and the probes use exact match (target and
type), which is the stricter criterion.

\paragraph{Interventions and action names.}
The tool-level action names are invented, like the rest of the
vocabulary, as an anti-prior precaution: \textsc{block} (suppress to
zero), \textsc{boost} (saturate), \textsc{dampen} (partial reduction),
plus signal activation. The paper's prose and figures write these as
block, increase, and decrease. Signals can be activated but not
inhibited, mirroring limited interventional access in real biology.

\paragraph{Non-LLM baselines.}
Three baselines calibrate the task (Figure~\ref{fig:gap}b).
\emph{Random} predicts 0.5 for every phenotype and runs no experiments;
it quantifies the structural floor of each score. \emph{Heuristic}
blocks each gene once and predicts by summing individual effects.
\emph{CausalSweep} is a calibration baseline with privileged access to
the sigmoid dynamics model (not to the network weights): it blocks and
boosts every gene systematically and predicts by inverting the sigmoid.
No LLM agent is given the underlying formula. CausalSweep's
near-perfect prediction accuracy with near-chance diagnosis
(Table~\ref{tab:gap-ci}) shows that exhaustive, well-calibrated
experimentation does not by itself produce causal identification.

\section{Supplementary methods and tables}
\label{app:methods}

\subsection{The gap, with confidence intervals}
\label{app:gap-ci}

Table~\ref{tab:gap-ci} gives the per-model scores behind
Figure~\ref{fig:gap}b, with 95\% confidence intervals
(t-distribution over 20 seeds, medium disease; missing scores count as
0), plus prediction accuracy on the two held-out tasks. Rescue exceeds
diagnosis for every model, and the two intervals fail to overlap for
six of the ten agents. The final row is a no-intervention
control: an empty intervention scored by the same metric. It is not
zero because in 3 of the 20 organisms the aberrant phenotype lies
within 0.02 of its healthy range, so simulation noise alone can move
it inside. The Random baseline scores no better than this control, so
random interventions essentially never rescue.

\begin{table}[!ht]
\centering
\caption{\textbf{Diagnosis, rescue, and prediction accuracy with 95\%
confidence intervals.} Medium disease, 20 seeds per agent, mean $\pm$
95\% CI. Seven LLMs, then the three non-LLM baselines.}
\label{tab:gap-ci}
\footnotesize
{\renewcommand{\arraystretch}{1.05}%
\begin{tabular}{lccc}
\toprule
agent & prediction acc. & diagnosis & rescue \\
\midrule
Opus 4.6      & $0.850 \pm 0.056$ & $0.550 \pm 0.239$ & $1.000 \pm 0.000$ \\
Sonnet 4.6    & $0.917 \pm 0.054$ & $0.350 \pm 0.229$ & $1.000 \pm 0.000$ \\
GPT-4.1       & $0.817 \pm 0.084$ & $0.250 \pm 0.208$ & $0.650 \pm 0.229$ \\
Qwen 3.5      & $0.792 \pm 0.066$ & $0.200 \pm 0.192$ & $0.900 \pm 0.144$ \\
DeepSeek V3   & $0.708 \pm 0.061$ & $0.200 \pm 0.192$ & $0.900 \pm 0.144$ \\
GPT-5.1       & $0.783 \pm 0.116$ & $0.150 \pm 0.171$ & $0.400 \pm 0.235$ \\
GPT-4o        & $0.642 \pm 0.068$ & $0.075 \pm 0.115$ & $0.100 \pm 0.144$ \\
\midrule
CausalSweep   & $0.933 \pm 0.047$ & $0.100 \pm 0.144$ & $0.650 \pm 0.229$ \\
Heuristic     & $0.625 \pm 0.061$ & $0.100 \pm 0.144$ & $0.700 \pm 0.220$ \\
Random        & $0.400 \pm 0.078$ & $0.075 \pm 0.115$ & $0.200 \pm 0.192$ \\
\midrule
No intervention & --- & --- & $0.150 \pm 0.171$ \\
\bottomrule
\end{tabular}}
\end{table}

\subsection{The discrimination census}
\label{app:census}

For each organism we enumerate every possible fault (agent-visible
genes $\times$ locked high/low, signals $\times$ chronic activation),
keep the \emph{alternatives} --- faults whose predicted baseline matches
the true baseline within $\pm 0.10$ --- and then simulate every available
intervention under each alternative. An intervention is
\emph{discriminating} for an alternative when its predicted outcomes
under the two candidate faults differ by more than $\pm 0.10$ on at
least one phenotype; the organism's discriminating set is the union
over its alternatives. Organisms with alternatives have 2.3 of them on
average, and a non-empty discriminating set holds on average 11.8 of
the 63 available interventions --- the one-in-five base rate that
Section~\ref{sec:discriminate} tests agents' choices against. Within the 39
resolvable organisms, 24 have a single experiment that separates the
truth from every alternative at once; the rest need one experiment per
alternative. Table~\ref{tab:census} gives diagnosis by ambiguity class
with the confidence intervals that Section~\ref{sec:evidence} states
qualitatively.

\begin{table}[!ht]
\centering
\caption{\textbf{Diagnosis accuracy by ambiguity class.} Bootstrap 95\%
CIs clustered by organism (organisms resampled, not sessions), since
sessions on one organism share a task.}
\label{tab:census}
\footnotesize
{\renewcommand{\arraystretch}{1.05}%
\begin{tabular}{lrrc}
\toprule
class & organisms & committed sessions & diagnosis [95\% CI] \\
\midrule
unambiguous           & 14 & 98  & 0.31 [0.16, 0.46] \\
resolvable ambiguity  & 39 & 273 & 0.30 [0.23, 0.37] \\
structural ambiguity  & 7  & 49  & 0.12 [0.00, 0.29] \\
\bottomrule
\end{tabular}}
\end{table}

\subsection{Exploration axes}
\label{app:explore}

Section~\ref{sec:explore} describes experimentation along three axes,
computed per session from the logged trajectories. \emph{Volume}: the
number of experiment calls (each costs one budget unit).
\emph{Breadth}: coverage, the fraction of the organism's genes probed
at least once (signals not counted). \emph{Direction}: a target is
tested bidirectionally when the same session pushes it both down and
up. Table~\ref{tab:explore} gives the per-model means over all three
disease sweeps. The diagnosis and rescue columns here use exact-match
scoring pooled over all 60 organisms, so they differ from
Table~\ref{tab:gap-ci} (graded scoring, medium sweep only).

\begin{table}[!ht]
\centering
\caption{\textbf{Exploration axes per model}, mean per session over all
420 sessions (60 per model). Diagnosis and rescue are exact-match,
pooled over the three disease sweeps.}
\label{tab:explore}
\footnotesize
{\renewcommand{\arraystretch}{1.05}%
\begin{tabular}{lrrrrrr}
\toprule
model & experiments & budget used & coverage & bidir.\ rate & diagnosis & rescue \\
\midrule
Opus 4.6      & 24.1 & 84\% & 1.00 & 0.20 & 0.53 & 0.98 \\
Sonnet 4.6    & 22.9 & 78\% & 1.00 & 0.26 & 0.40 & 0.94 \\
Qwen 3.5      & 21.5 & 75\% & 0.83 & 0.36 & 0.30 & 0.94 \\
DeepSeek V3   & 18.3 & 67\% & 0.52 & 0.43 & 0.22 & 0.93 \\
GPT-5.1       & 17.1 & 59\% & 0.91 & 0.23 & 0.15 & 0.46 \\
GPT-4.1       & 12.7 & 47\% & 0.40 & 0.37 & 0.28 & 0.73 \\
GPT-4o        & 7.4  & 28\% & 0.26 & 0.23 & 0.07 & 0.61 \\
\bottomrule
\end{tabular}}
\end{table}

\subsection{Probe construction and full probe tables}
\label{app:probes}

\paragraph{Probes 1--2 (isolated evidence).}
Single-turn items built from the census and the session logs: 69
\emph{real} items (a discriminating result; predicted separations
$\geq 0.14$) and 43 \emph{fillers} (an uninformative result; every
phenotype gap $\leq 0.03$), 112 items in total, 96 of whose observed
values come from real session logs. Probe~1 shows the two candidate
faults, their predicted readouts, and the observed value, and asks
which fault the result supports (A / B / UNDETERMINED). Probe~2 adds an
assigned working hypothesis and asks whether to retain or revise it.
All seven models ran at a uniform 512-token response budget.
Verbatim prompts for all probes, and the session task instructions,
are in Appendix~\ref{app:prompts}.
Uncertainty is $\pm$ half the 95\% Wilson interval; cells at a boundary
show the one-sided bound. A full second run of probes 1--2 (all seven
models) moves the 35 table cells by 0.025 on average; real-item cells
are essentially exact, filler cells move by up to 0.14 --- filler
differences under about 0.10 between models should not be interpreted.

\begin{table}[!ht]
\centering
\caption{\textbf{Probes 1--2: item accuracy per model.} Real items
$n{=}69$ (revise 34, retain 35), fillers $n{=}43$. Filler accuracy
means answering UNDETERMINED; \emph{revise-on-noise} counts filler
items answered REVISE --- switching hypothesis on an uninformative
result.}
\label{tab:probes12}
\footnotesize
{\renewcommand{\arraystretch}{1.05}%
\begin{tabular}{lcccccc}
\toprule
model & P1 real & P1 filler & P2 revise & P2 retain & P2 filler & revise-on-noise \\
\midrule
Opus 4.6      & 1.00 ($\geq$0.95) & $0.35 \pm 0.14$ & 1.00 ($\geq$0.90) & 1.00 ($\geq$0.90) & $0.63 \pm 0.14$ & 5/43 \\
Sonnet 4.6    & 1.00 ($\geq$0.95) & $0.44 \pm 0.14$ & 1.00 ($\geq$0.90) & 1.00 ($\geq$0.90) & $0.51 \pm 0.14$ & 5/43 \\
GPT-4.1       & 1.00 ($\geq$0.95) & $0.33 \pm 0.13$ & 1.00 ($\geq$0.90) & $0.97 \pm 0.07$ & $0.63 \pm 0.14$ & 5/43 \\
GPT-4o        & 1.00 ($\geq$0.95) & $0.30 \pm 0.13$ & 1.00 ($\geq$0.90) & $0.94 \pm 0.09$ & $0.40 \pm 0.14$ & 13/43 \\
GPT-5.1       & 1.00 ($\geq$0.95) & $0.53 \pm 0.14$ & 1.00 ($\geq$0.90) & 1.00 ($\geq$0.90) & $0.35 \pm 0.14$ & 8/43 \\
DeepSeek V3   & $0.97 \pm 0.05$ & $0.37 \pm 0.14$ & $0.94 \pm 0.09$ & $0.74 \pm 0.14$ & $0.58 \pm 0.14$ & 10/43 \\
Qwen 3.5      & 1.00 ($\geq$0.95) & $0.33 \pm 0.13$ & 1.00 ($\geq$0.90) & 1.00 ($\geq$0.90) & $0.60 \pm 0.14$ & 5/43 \\
\bottomrule
\end{tabular}}
\end{table}

\paragraph{Probe 3 (evidence in context).}
Fixed replay: for each of the 43 organisms with a resolvable
alternative, one real session's full pre-commit evidence (median 22
experiments; 42 of the 43 records come from Opus 4.6 sessions
and one from a Sonnet 4.6 session) is
rendered as a neutral log. All models read the same 43 logs, each as a single-turn
question, at a 1024-token response budget. The three conditions remove the scaffolding
one step at a time. Probe~3a gives the two candidate faults and
the oracle's predicted readout of all three phenotypes under each
candidate, for every experiment in the log, followed by Probe~1's noise
sentence. Probe~3b withholds those predictions and gives the
two candidate faults alone (A / B / UNDETERMINED); probe~3c
asks openly for the fault (target and type, exact match). The live
column repeats each model's live exact-match diagnosis on the same 43
organisms. Models answered UNDETERMINED in 1 of 301 3b answers --- the
pick-a-side habit of Table~\ref{tab:probes12} persists in context.

A source-robustness run repeats Probes~3a and~3b on records
written by Qwen 3.5 instead: the same 43 organisms, the same competing
causes and assignment, logs of the same median length, 602 calls with
no errors. Accuracy is 0.84--0.95 with the predictions supplied against
0.37--0.72 without them, a median gain of 0.35 against 0.35 on the
Opus-sourced records; every model improves and all seven interval pairs
are disjoint on both sets. Per-model differences between the two source
sets are within their intervals and are not interpreted.

Probe~3a exists because Probe~3b differs from Probe~1 in four
ways at once: the relevant experiment is no longer named, the relevant
phenotype is no longer named, the per-candidate predictions are no
longer given, and the noise tolerance is no longer stated. Probe~3a
restores the last two while leaving the decisive observation buried,
and reuses the Probe~3b items unchanged --- same organisms, same source
sessions, same transcripts, same candidate assignment. The untreated baseline is deliberately not a row: an alias is
defined as matching the true fault at baseline within the $0.10$
separation threshold, so the residual gap (median $0.057$ over the 43
items) is not evidence by that criterion, yet is large enough to answer
from without reading the log. Every item retains at least one row whose
two predictions separate by at least $0.10$; across those rows 174
favour the true fault and 6 favour the alias, all 6 on the hard tier.
Pooled accuracy by difficulty is 0.98 (medium), 0.96 (easy) and 0.89
(hard), with 14 errors in 301 answers and no unparsed or UNDETERMINED
reply.

\begin{table}[!ht]
\centering
\caption{\textbf{Probe 3: the same evidence, inside an investigation
record.} $n{=}43$ per cell, $\pm$ 95\% Wilson half-widths; cells at the
ceiling show the one-sided bound. Probe~3a supplies the
candidate predictions for every logged experiment, 3b withholds them.
The replayed logs come almost entirely from Opus 4.6
sessions, so 3c
measures reasoning over that evidence, not what each model's own
investigation would have produced.}
\label{tab:probe3}
\footnotesize
{\renewcommand{\arraystretch}{1.05}%
\begin{tabular}{lcccc}
\toprule
model & 3a (predictions given) & 3b (predictions derived) & 3c (open diagnosis) & live \\
\midrule
Opus 4.6      & 1.00 ($\geq$0.92) & $0.65 \pm 0.14$ & $0.58 \pm 0.14$ & $0.58 \pm 0.14$ \\
Sonnet 4.6    & 1.00 ($\geq$0.92) & $0.58 \pm 0.14$ & $0.47 \pm 0.14$ & $0.44 \pm 0.14$ \\
GPT-4.1       & $0.95 \pm 0.07$ & $0.70 \pm 0.13$ & $0.49 \pm 0.14$ & $0.28 \pm 0.13$ \\
GPT-4o        & $0.98 \pm 0.06$ & $0.74 \pm 0.13$ & $0.37 \pm 0.14$ & $0.09 \pm 0.09$ \\
GPT-5.1       & $0.91 \pm 0.09$ & $0.56 \pm 0.14$ & $0.42 \pm 0.14$ & $0.12 \pm 0.10$ \\
DeepSeek V3   & $0.95 \pm 0.07$ & $0.51 \pm 0.14$ & $0.51 \pm 0.14$ & $0.16 \pm 0.11$ \\
Qwen 3.5      & $0.88 \pm 0.10$ & $0.53 \pm 0.14$ & $0.77 \pm 0.12$ & $0.26 \pm 0.13$ \\
\bottomrule
\end{tabular}}
\end{table}

\subsection{The full case-study trace}
\label{app:trace}

\begin{figure}[!ht]
\centering
\includegraphics{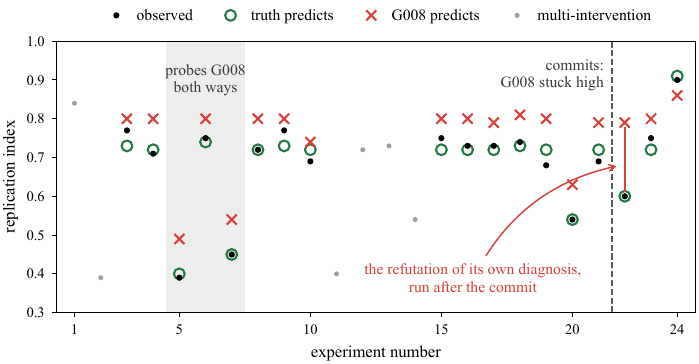}
\caption{\textbf{The full session record behind
Figure~\ref{fig:trace}.} Observed readouts (black) against the
predictions of the true fault (green rings) and of the agent's eventual
diagnosis, G008 (red crosses); gray dots are multi-intervention calls,
for which no single-fault prediction is defined. The shaded band marks
the three G008 probes of Figure~\ref{fig:trace}; the dashed line marks
the commit; the red segment at experiment 22 is the post-commit
refutation.}
\label{fig:trace-full}
\end{figure}

Figure~\ref{fig:trace-full} shows the complete 24-experiment record of
the session excerpted in Figure~\ref{fig:trace}, including the
multi-intervention calls the excerpt elides. Two details complete the
story told in Section~\ref{sec:discriminate}. First, none of the agent's
solo pre-commit experiments formally separates the truth from its
eventual diagnosis at the $\pm 0.10$ criterion --- the readouts
consistently favour the truth without being decisive, which is why the
figure says \emph{favours}. Second, at experiment 22, after the commit
locked its hypothesis, the agent ran one of the four experiments that
do formally separate the two candidates: the observed 0.60 against the
0.79 its diagnosis predicts is the decisive refutation of its own
committed hypothesis, run too late to matter.

\subsection{Task instructions and probe prompts}
\label{app:prompts}

The excerpts below are verbatim from the benchmark and probe code;
only line wrapping is adjusted, and \promptvar{angle brackets} mark
values filled in per item. BLOCK / BOOST / DAMPEN are the invented
tool-level action names discussed in Appendix~\ref{app:env}; the
paper's prose writes them as block / increase / decrease. The complete
session system prompt and tool schemas are in the released code.

\paragraph{Session system prompt (excerpts).}
Every session opens with the anti-prior block:
\begin{defbox}\small
CRITICAL: All gene names (e.g. G001, G042), signal labels (e.g.
``nutrient-X''), and phenotype names (e.g. ``growth-rate'') are
INVENTED VOCABULARY. They do NOT correspond to any real biological
system. Prior biological knowledge is IRRELEVANT and may mislead you.
The only valid path to understanding this organism is systematic
causal reasoning from perturbation experiments.
\end{defbox}
Disease sessions add:
\begin{defbox}\small
DISEASE CONTEXT: This cell exhibits aberrant phenotypes compared to
its healthy baseline ranges. The regulatory network has a single
malfunction --- either a node locked high (LOCK\_HI), a node locked
low (LOCK\_LO), or a persistently active external signal (PERSIST).
Your job is to identify it and restore healthy phenotypes.
\end{defbox}
The diagnosis and rescue tasks, from the prompt's PRIMARY GOALS list:
\begin{defbox}\small
1. Malfunction diagnosis --- identify which gene or signal is broken
and how (LOCK\_HI / LOCK\_LO / PERSIST). Submit via
submit\_malfunction\_hypothesis().

2. Rescue --- design an intervention that restores aberrant phenotypes
to healthy ranges. Test it via test\_rescue().
\end{defbox}
and the workflow step that makes the commitment binding:
\begin{defbox}\small
When confident, submit your malfunction hypothesis via
submit\_malfunction\_hypothesis(). This is a one-time, irreversible
commitment --- it locks immediately and unlocks test\_rescue(). Score
is revealed at session end.
\end{defbox}

\paragraph{Tool descriptions.}
The two disease-mode tools, as described to the agent:
\begin{defbox}\small
\texttt{submit\_malfunction\_hypothesis} --- ``Submit your hypothesis
about which gene or signal is malfunctioning and how. No budget cost.
One submission only --- locked immediately. Unlocks test\_rescue().
Score is NOT returned here; it is revealed at session end. Commit only
when confident --- you cannot revise this.'' Arguments:
\texttt{target\_name} (a gene name or signal label) and
\texttt{malfunction\_type} (LOCK\_HI, LOCK\_LO, or PERSIST).

\texttt{test\_rescue} --- ``Test whether a proposed intervention
restores aberrant phenotypes to healthy ranges. Costs 1 budget.
Requires submit\_malfunction\_hypothesis() to be called first. Runs
the organism from the disease baseline with your intervention and
reports per-phenotype restoration status and MSE vs healthy
baseline.'' Argument: a map of targets to interventions (BLOCK /
BOOST / DAMPEN for genes, or signal activation).
\end{defbox}

\paragraph{Probe 1 (read one clue).}
The full single-turn prompt:
\begin{defbox}\small\obeylines
You are analysing a synthetic organism with exactly one hidden fault.
The fault is one of these two candidates:
\hspace*{1em}(A) \promptvar{candidate A}
\hspace*{1em}(B) \promptvar{candidate B}
~
An experiment was run: \promptvar{experiment}.
Predicted `\promptvar{phenotype}' readout:
\hspace*{1em}if (A) is true: about \promptvar{value}
\hspace*{1em}if (B) is true: about \promptvar{value}
Observed `\promptvar{phenotype}': \promptvar{value}
Predictions and readouts carry simulation noise; deviations up to
about $\pm$0.05 from a matching prediction are common.
~
Which candidate does this result support?
Answer with exactly one token: A, B, or UNDETERMINED.
\end{defbox}
Candidates render as, e.g., ``gene G012 is locked high (LOCK\_HI)'' or
``signal burst-Y is persistently active (PERSIST)''; experiments as
``apply BLOCK to gene G012'' or ``activate signal nutrient-X''.

\paragraph{Probe 2 (update a belief).}
Identical to Probe~1, with one line inserted after the candidates ---
\begin{defbox}\small
Your current working hypothesis is (\promptvar{A or B}).
\end{defbox}
--- and the closing question replaced by:
\begin{defbox}\small
Given this result, should you RETAIN your hypothesis, REVISE it to the
other candidate, or is the result UNINFORMATIVE between the two?

Answer with exactly one token: RETAIN, REVISE, or UNINFORMATIVE.
\end{defbox}

\paragraph{Probe 3 (diagnose from the log).}
Each item replays one session's pre-diagnosis experiment log. The
transcript format:
\begin{defbox}\small\obeylines
A budget-limited investigation was run on a synthetic organism with
exactly one hidden fault (a gene locked high or low, or a signal
persistently active).
Genes: \promptvar{gene list}
Signals: \promptvar{signal list, with modes}
Baseline phenotypes (diseased) vs healthy ranges:
\hspace*{1em}\promptvar{label}: \promptvar{value} (healthy \promptvar{lo}--\promptvar{hi}) \texttt{<-{}- ABERRANT} \textit{[if outside]}
\hspace*{1em}\ldots
Experiments run, in order (each line: interventions \texttt{->} resulting phenotypes):
\hspace*{1em}1. \promptvar{action} \promptvar{target} \texttt{->} \promptvar{phenotype values}
\hspace*{1em}\ldots
\end{defbox}
Probe 3b appends the two candidates and asks:
\begin{defbox}\small\obeylines
Exactly one of these two candidate faults is present:
\hspace*{1em}(A) \promptvar{candidate A}
\hspace*{1em}(B) \promptvar{candidate B}
~
Based on the evidence above, which candidate is more likely?
You may reason briefly first. End your reply with one line:
ANSWER: A\hspace*{2em}or\hspace*{2em}ANSWER: B\hspace*{2em}or\hspace*{2em}ANSWER: UNDETERMINED
\end{defbox}
Probe 3c removes the candidates and asks for an open diagnosis:
\begin{defbox}\small\obeylines
Based on the evidence above, name the single most likely fault.
Fault types: LOCK\_HI, LOCK\_LO, PERSIST. You may reason briefly
first. End your reply with one line:
ANSWER: \promptvar{target} \promptvar{type}\hspace*{2em}e.g.\hspace*{2em}ANSWER: G007 LOCK\_HI
\end{defbox}

\paragraph{Probe 3a (predictions supplied).}
Probe 3b's prompt, with one block inserted between the
candidates and the question:
\begin{defbox}\small\obeylines
Predicted readouts under each candidate, for every experiment above:
\hspace*{1em}\promptvar{n}. \promptvar{interventions}
\hspace*{2.5em}if (A) is true: \promptvar{phenotype readouts}
\hspace*{2.5em}if (B) is true: \promptvar{phenotype readouts}
\hspace*{1em}\ldots
~
Predictions and readouts carry simulation noise; deviations up to
about $\pm$0.05 from a matching prediction are common.
\end{defbox}

\end{document}